\documentclass{INTERSPEECH2023}

\interspeechcameraready

\title{Let's Give a Voice to Conversational Agents in Virtual Reality}
\name{Michele Yin, Gabriel Roccabruna, Abhinav Azad, Giuseppe Riccardi}
\address{
  Signals and Interactive Systems Lab, University of Trento, Italy}
\email{gabriel.roccabruna@unitn.it, giusppe.riccardi@unitn.it}

\begin{document}
\maketitle
\begin{abstract}
The dialogue experience with conversational agents can be greatly enhanced  with multimodal and immersive interactions in virtual reality.
In this work, we present an open-source architecture with the goal of simplifying the development of conversational agents operating in virtual environments. The architecture offers the possibility of plugging in conversational agents of different domains and adding custom or cloud-based Speech-To-Text and Text-To-Speech models to make the interaction voice-based. Using this architecture, we present two conversational prototypes operating in the digital health domain developed in Unity for both non-immersive displays and VR headsets. 
The architecture is publicly available on GitHub \footnote{GitHub repo \url{https://github.com/sislab-unitn/Let-s-Give-a-Voice-to-Conversational-Agents-in-VR}}.

\end{abstract}

\section{Introduction}
Extended reality, namely virtual, augmented and mixed realities, has been gaining attention and interest in the research and industry sectors. Virtual Reality (VR) allows the development of interactive Virtual Environments (VE), which can be a surrogate rendering of imaginary or real scenes. VEs can be displayed on non-immersive screens, such as a TV screen, or on head-mounted displays designed for VR, called VR headsets. The advantage of these headsets with respect to non-immersive devices is the enhanced immersive experience by giving the feeling to the user of physically entering into the virtual world and, through haptic controllers, interacting with it in a quasi-natural way. One of the applications of this technology is the training of humans in different skills by simulating working environments such as a surgery room, firestorm or a classroom \cite{kaplan2021effects}. In these simulations, Embodied Conversational Agents (ECA), which are multimodal Conversational Agents (CA) with a 3D human-like look, are crucial in assisting the users during the training process \cite{macias2020carla} or in acting in the role of other human counterparts \cite{sonlu2021conversational}. In these applications, the interaction between an ECA and a human counterpart has to be spoken to be effective. However, most of the CAs can communicate via text only.

In this work, we present an open-source architecture to interface CAs to VEs. With this architecture, CAs can hold voice-based conversations with a user in a VE by leveraging two customisable modules for Speech To Text (STT) and Text To Speech (TTS). We test our architecture on a simulation of a point of care in which we designed and developed two CAs with the tasks of collecting symptoms (triage room) and the patient's medical history (anamnesis room). In this project, we used Unity to develop the VE designed to run on Oculus Quest 2 VR headsets\footnote{\url{https://www.meta.com/us/quest/products/quest-2/}}, Rasa\footnote{\url{https://rasa.com}} to develop the CA, publicly available avatars\footnote{\url{https://readyplayer.me}} and, publicly available STT\footnote{\url{https://huggingface.co/speechbrain/asr-wav2vec2-librispeech}} and TTS\footnote{\url{https://huggingface.co/microsoft/speecht5_tts}} models.

\section{Architecture}

\begin{figure}[t]
	\begin{center}
	   \includegraphics[scale=0.53]{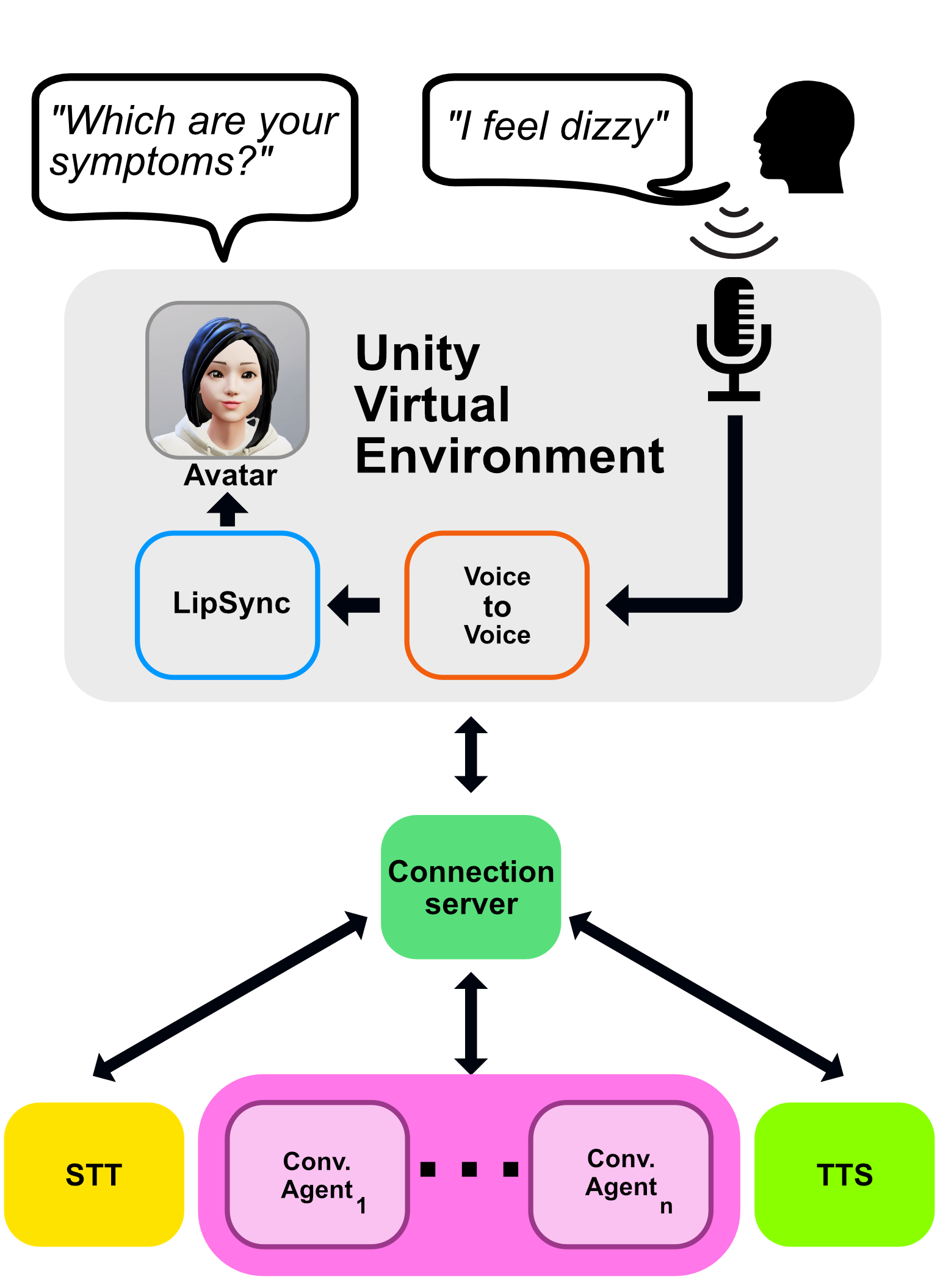}
	\end{center}
	\caption{Schema of our Architecture}
	\label{fig:architecture}
\end{figure}

The architecture is composed of 5 main modules presented in Figure \ref{fig:architecture}, which are Unity Virtual Environment (UVE), connection server, Conversational Agent, Speech-To-Text (STT) and Text-To-Speech (TTS). 

UVE represents the virtual scene where the user can interact with the simulated environment and communicate via voice with the avatar. UVE is composed of two sub-modules namely, LipSync and Voice To Voice. LipSync is used to mimic the human's facial movements, occurring while speaking, on the avatar. Voice To Voice is in charge of exchanging the audio files of the user's utterances, recorded in the UVE, and CA's utterances, played by the LipSync, with the connection server. The Connection Server (CS) is the hub of the architecture, which connects the CAs to the UVE and all the other modules.
The CS takes as input the audio file of the user's utterance recorded in UVE. Since the CA works with text only, the CS gets the transcription of the user's utterance by sending a request to the STT module. The STT and TTS modules are model agnostic meaning that they work with any model which can be either cloud-based or running locally. The resulting transcription is then sent to the CA which yields a coherent text-based response. The CA's response is routed back to the CS which gets its speech synthesis by querying the TTS module. At this stage, the synthesised response of the CA is sent to the UVE and played by the LipSync module. 

Our architecture can handle different CAs at the same time, the selection of the CAs is decided by the UVE. Furthermore, the UVE can also decide the voice tone of the avatar based on the options provided by the TTS.

The CA's response time, along with the content, is an important aspect to maximise the user's experience. In human-human conversations, the time elapsed between two turns, i.e. the response time, is usually within 500 milliseconds \cite{strombergsson13_interspeech}. To minimise the difference between this threshold and CA's response, we selected the Wav2Vec2.0 for STT, with an average processing time of 0.8 seconds on CPU, and SpeechT5 for TTS. However, the TTS is twice as slow as the STT, i.e. 1.7 seconds on CPU\footnote{STT and TTS processing times have been computed on the same sentences coming from our use-case scenario.}.  To mitigate this issue, we hide the processing time of the TTS in the natural pauses of the dialogue. In particular, we chunk the CA's responses based on the punctuation. The resulting chunks are then synthesised sequentially and played as soon as the audio file of a chunk is received by the UVE. Furthermore, the processing time is further masked by the fact that the processing time of the TTS is always lower than the duration of the corresponding audio, therefore, while the avatar is playing the first chuck the other chunks are synthesised and ready to be played. Nevertheless, to exploit this advantage the lengths in terms of tokens of the chunks have to be approximately the same, which can be achieved by adding strategic punctuation marks.

\section{Scenario}
\begin{figure}[t]
	\begin{center}
	   \includegraphics[scale=0.18]{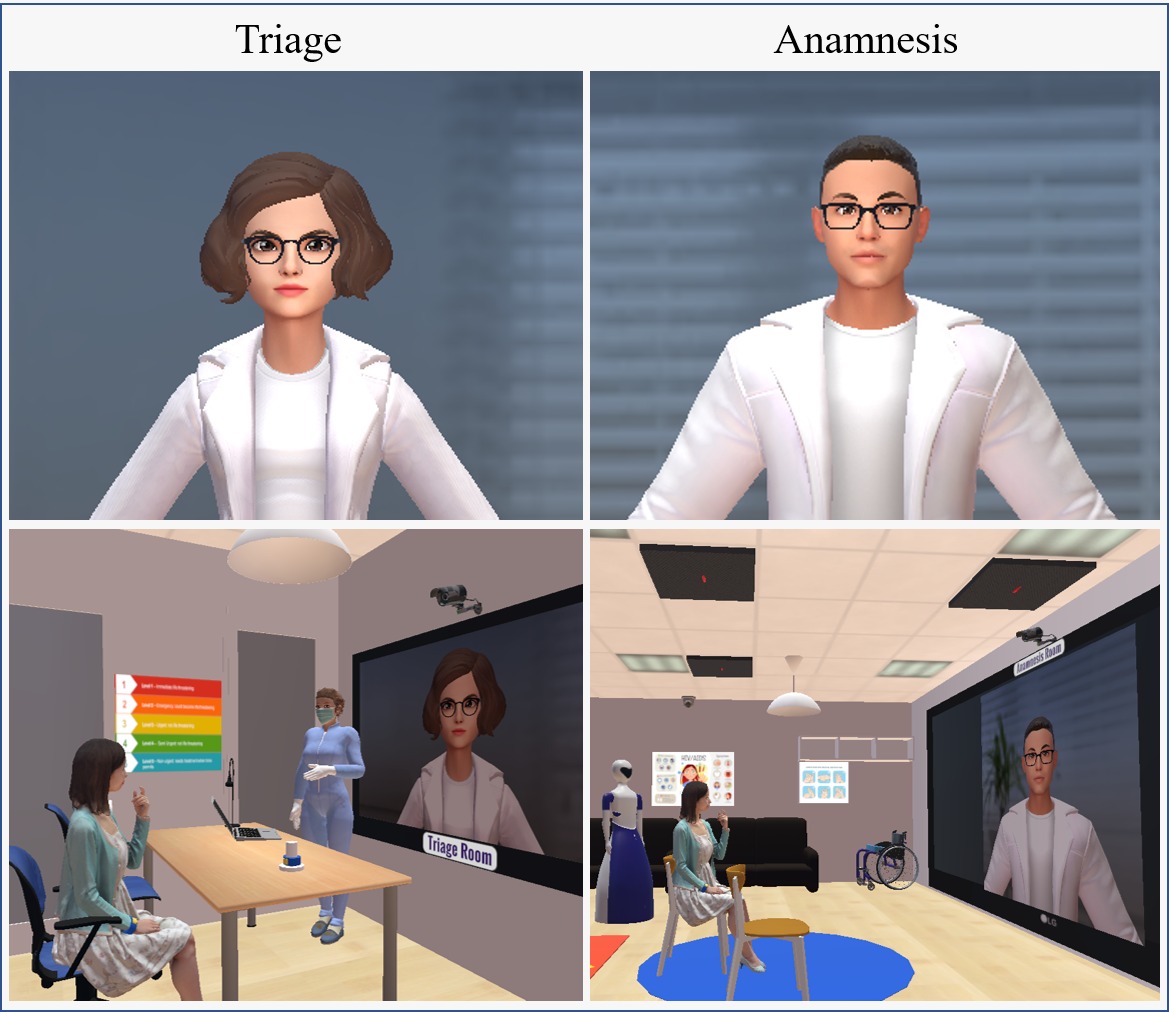}
	\end{center}
	\caption{The figure shows avatars on the top and virtual rooms at the bottom of the triage and anamnesis rooms.}
	\label{fig:scene}
\end{figure}

We tested our architecture for a health application in a virtual point-of-care simulation. The virtual point of care is composed of triage and anamnesis rooms each of them equipped with an avatar linked to a conversational agent. In the triage room, the CA asks the patient a series of questions to collect the required information to assess the patient's priority level based on the severity of the symptoms expressed with five colour codes (cyan, green, yellow, orange and red). In the anamnesis room, the CA checks again the patient's symptoms and collects information about the patient's medical history. The virtual scenes of the two rooms and the corresponding avatars are depicted in Figure \ref{fig:scene}.
\section{Conclusions and Further directions}
This paper presents an open-source architecture to integrate conversational agents into a virtual environment tested on a simulation of a virtual point of care. Our future directions are to integrate body gestures, managed by the conversational agent, and ground the conversation based on the elements displayed in the virtual environment. Furthermore, we plan to estimate the stress level that the user perceives by analysing the speech signal \cite{stepanov2018depression} and/or biomarkers (e.g. heart rate or electrodermal activity) \cite{ghosh2015annotation} measured by wearable devices. We believe that this architecture could be also useful in the design of validation protocols to test STT and TTS models in the field. 

\bibliographystyle{IEEEtran}
\bibliography{mybib}

\end{document}